\documentclass[letterpaper, 10 pt, conference]{ieeeconf}  % Comment this line out if you need a4paper

\IEEEoverridecommandlockouts                              % This command is only needed if
\usepackage[caption=false,font=footnotesize]{subfig}

\usepackage{svg}
\usepackage{cite}
\usepackage{amsmath}
\usepackage{todonotes}
\usepackage{soul}
\usepackage{tikz}
\usepackage{multirow}
\usepackage{changes}
\usetikzlibrary{shapes,arrows,fit,backgrounds}
\pgfdeclarelayer{bg}    % declare background layer
\pgfsetlayers{bg,main}  % set the order of the layers (main is the standard layer)

\makeatletter\if@todonotes@disabled\else\fi\makeatother

\DeclareMathOperator*{\argmin}{arg\,min}

\title{\LARGE \bf
 Multi-Modal Loop Closing in Unstructured Planetary Environments with Visually Enriched Submaps
}

\author{Riccardo Giubilato, Mallikarjuna Vayugundla, Wolfgang St\"urzl, Martin J. Schuster, \\Armin Wedler, Rudolph Triebel% <-this % stops a space
\thanks{All authors are with German Aerospace Center (DLR), Institute of Robotics and Mechatronics, M{\"u}nchener Str. 20, 82234, Wessling, Germany {\tt\small \{firstname.lastname@dlr.de\}}
}}

\begin{document}
\bstctlcite{BibControl}
\maketitle

\thispagestyle{empty}
\pagestyle{empty}
\begin{abstract}
Future planetary missions will rely on rovers that can autonomously explore and navigate in unstructured environments. An essential element is the ability to recognize places that were already visited or mapped.
In this work we leverage the ability of stereo cameras to provide both visual and depth information, guiding the search and validation of loop closures from a multi-modal perspective. We propose to augment submaps that are created by aggregating stereo point clouds, with visual keyframes. Point clouds matches are found by comparing CSHOT descriptors and validated by clustering while visual matches are established by comparing keyframes using Bag-of-Words (BoW) and ORB descriptors. The relative transformations resulting from both keyframe and point cloud matches are then fused to provide pose constraints between submaps in our graph-based SLAM framework. Using the LRU rover, we performed several tests in both indoor laboratory environment as well as challenging planetary analog environment on Mount Etna, Italy, consisting of areas where either keyframes or point clouds alone failed to provide adequate matches demonstrating the benefit of the proposed multi-modal approach.
\end{abstract}

\section{Introduction}
Simultaneous Localization and Mapping (SLAM) enables autonomous robots to explore unknown and GPS-denied environments. The ability to create drift-free maps in challenging outdoor environments is crucial for the accurate localization of scientifically relevant targets or for search-and-rescue related tasks. In the context of space exploration, stereo cameras are widely used as a mechanically simple sensor, delivering both visual and depth information about the observed environment.
A major advantage of depth information is that up to a certain degree the 3D structure of an environment can be estimated independent of light conditions and viewpoints.
On the other hand, areas without suitable 3D features might still provide meaningful visual cues. In addition, the range for obtaining reliable depth information, in particular from stereo vision is limited, as accuracy and resolution decreases quickly with distance.
Therefore, in order to obtain sufficient information for loop closure detection and validation in challenging environments it is advantageous to exploit both modalities.

 In this paper we present the following major contributions:
 \begin{itemize}
 	\item We propose a loop closure framework which leverages both the 3D structure and the visual appearance provided by stereo cameras in the context of a submap-based SLAM system, allowing to maximize pose accuracy and to increase the chance of establishing inter-pose constraints.
 	\item We test the proposed framework both in an indoor laboratory environment comprising replicas of natural features and in an outdoor environment on Mount Etna, Italy, designated as an analog environment for lunar scenarios.
 \end{itemize}
 The paper is organized as follows: In section~\ref{sec:rel} we give an overview of the related work. After describing our submap-based SLAM pipeline (section~\ref{sec:slam}) we then, in section~\ref{sec:match}, present our approach for establishing multi-modal loop closures. In section~\ref{sec:exp} we evaluate the proposed system in several experiments and, finally, in section~\ref{sec:concl} we draw some conclusions.

% \begin{itemize}
%     \item general intro on SLAM - usual stuff
%     \item stereo camera as space-qualified sensors?
%     \item why it makes sense to use both stereo images and pointclouds! + remarks on the difference between LiDAR pointcloud - based SLAM and our specific scenario: buildings constrain well the problem of aligning clouds in city landscapes. In our case the same methods can not be used. + boundary of submaps depend on the trigger location and rover path, while lidar clouds are usually 360 $\rightarrow$ lidar clouds captured from the same pose will look identical, submaps started from the same pose will not $\implies$ no global descriptors
% \end{itemize}
\begin{figure}[t]
    \centering
    \includegraphics[width=\linewidth, trim=0 0.2cm 0 0, clip]{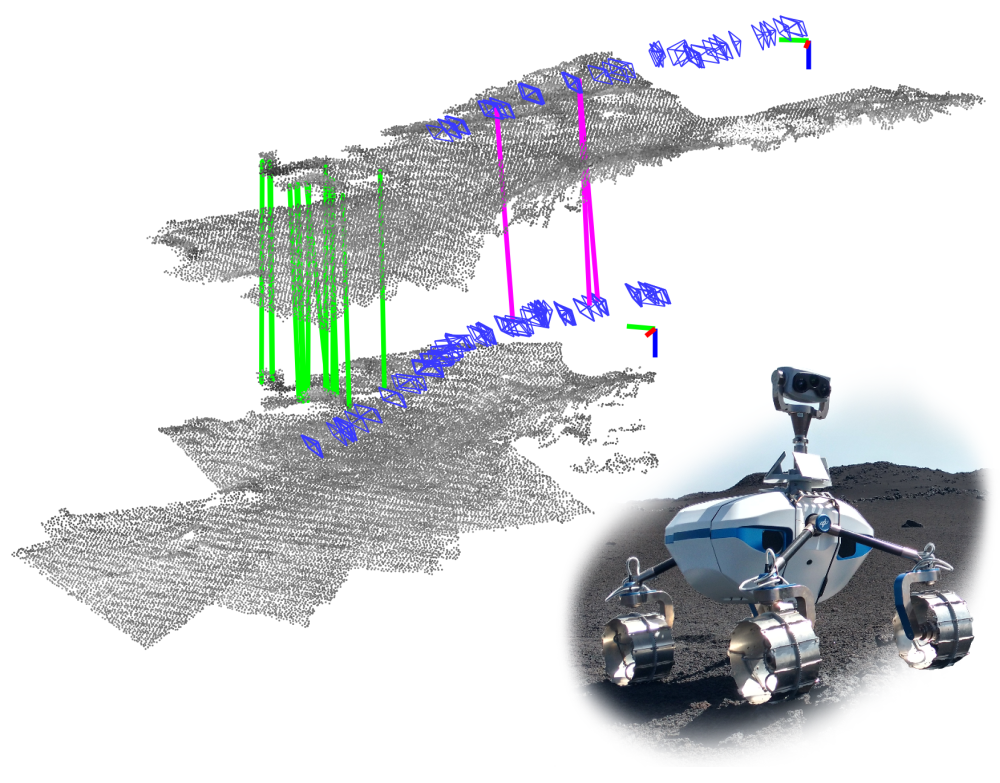}
    \caption{Example of matching submaps from the ``\textit{Etna}" sequences using 3D descriptor and keyframe correspondences. Green lines denote matching 3D keypoints validated after Hough3D clustering, magenta lines connect matching keyframes. The two reference systems represent the origins of both submaps (Submaps are displaced in the vertical direction for visualization purposes). Bottom: LRU rover on Mount Etna, a designated Moon-analog test site in Sicily, Italy.}
    \label{fig:my_label}
\end{figure}
\renewcommand{\sfdefault}{lmss}
\begin{figure*}[t]
	\centering
	\tikzstyle{decision} = [diamond, draw, aspect=1, fill=green!20,
	text width=8em, text badly centered, node distance=3cm, inner sep=2pt]
	\tikzstyle{blockFE} = [draw, rectangle, rounded corners,fill=green!20, node distance=2cm,
	minimum height=1em , text centered,
	text width=7em,]
	\tikzstyle{blockQU} = [draw, rectangle, rounded corners,fill=blue!20, node distance=2cm,
	minimum height=1em , text centered,
	text width=6em,]
	\tikzstyle{blockMA} = [draw, rectangle, rounded corners,fill=orange!20, node distance=2cm,
	minimum height=1em , text centered,
	text width=6em,]
	\tikzstyle{blockBE} = [draw, rectangle, rounded corners,fill=red!20, node distance=2cm,
	minimum height=1em , text centered,
	text width=5em,]
	\tikzstyle{empty} = [circle, text width=2.5em, inner sep=0pt, minimum height=1em]
	\tikzstyle{blockC} = [draw, rectangle, rounded corners,fill=black!10, node distance=2cm,
	minimum height=1em , text centered,
	text width=6em,]
	\tikzstyle{blockOT} = [draw, rectangle, rounded corners,fill=teal!20, node distance=2cm,
	minimum height=1em , text centered,
	text width=6em,]
	\tikzstyle{line} = [draw, -latex']
	\tikzstyle{line-short} = [draw, -latex', shorten >=0.5cm]
	\tikzstyle{line-sshort} = [draw, -latex', shorten >=0.1cm]
	\tikzstyle{cloud} = [draw, rectangle, rounded corners,fill=red!20, node distance=2cm,
	minimum height=1em , text centered,
	text width=6em,]
	\begin{tikzpicture}[node distance = 1cm, auto, font=\sf]

	% front end
	\node [cloud, label=below:{}] (sc) {\small Stereo Camera};
	\node [cloud, below of=sc, node distance=0.6cm, label=below:{}] (imu) {\small IMU};
	\node [blockFE, right of=sc, node distance=3.2cm] (clouds) {\small Aggregate clouds};
	\node [blockFE, below of=clouds, node distance=1.5cm] (orb) {\small ORB keypoints};
	\node [blockFE, below of=orb, node distance=0.7cm] (bucket) {\small Feature bucketing};
	\node [blockFE, below of=bucket, node distance=0.85cm] (msg) {\small Build ROS \texttt{submap\_msg}};

	% calibration
	\node[blockC, below of=imu, node distance=1.5cm] (sgbm) {\small SGM};
	\node[blockC, below of=sgbm, node distance=0.85cm] (vio) {\small Visual-Inertial Odometry};

	\begin{pgfonlayer}{bg}
	\node[draw=gray,dashed,fill=green!5,line width=1.2pt, rounded corners, label={\small \textbf{Submap Generation}}, fit=(clouds) (orb) (bucket) (msg)](submapgen) {};
	\node[line width=0pt, rounded corners, label={\small \textbf{Is Keyframe?}}, fit=(orb) (bucket) (msg)](subgen) {};
	\end{pgfonlayer}

	\begin{pgfonlayer}{bg}
	\node[draw=gray,dashed,fill=red!5,line width=1.2pt, rounded corners,  label={\small \textbf{Sensor Inputs}}, fit=(sc) (imu)](in) {};
	\end{pgfonlayer}
	\begin{pgfonlayer}{bg}
	\node[draw=gray,dashed,fill=black!5,line width=1.2pt, rounded corners,  label={\small \textbf{Front-End}}, fit=(sgbm) (vio)](fe) {};
	\end{pgfonlayer}

	% Queue
	\node [blockQU, below right = -1cm and 0.65cm of orb] (overlap) {\small Compute spatial overlap};
	\node [blockQU, below of=overlap, node distance=1.1cm] (cand) {\small Push candidates};
	\begin{pgfonlayer}{bg}
	\node[draw=gray,dashed,fill=blue!5,line width=1.2pt, rounded corners, label={\small \textbf{Priority Queue}}, fit=(overlap) (cand)](pq) {};
	\end{pgfonlayer}

	% Match
	\node [empty, right of=clouds, node distance=6.2cm] (emp) {};
	\node [blockMA, below of=emp, node distance=0.5cm] (kpts) {\small 3D keypoints};
	\node [blockMA, below of=kpts, node distance=0.9cm] (btex) {\small Compute and match CSHOT};
	\node [blockMA, below of=btex, node distance=1.3cm] (valid) {\small Validation: Hough3D + ICP};

	\node [empty, right of=emp, node distance=2.5cm] (emp2) {};
	\node [blockMA, below of=emp2, node distance=0.5cm] (bow) {\small BoW scores};
	\node [blockMA, below of=bow, node distance=0.6cm] (orbm) {\small Match ORB};
	\node [blockMA, below of=orbm, node distance=0.8cm] (pnp) {\small PnP + RANSAC};
	\node [blockMA, below of=pnp, node distance=1.05cm] (vvalid) {\small GN opt. + Validation};
	\begin{pgfonlayer}{bg}
	\node[draw=gray,dashed,fill=orange!5,line width=1.2pt, rounded corners, label={\small \textbf{Submap Matching}}, fit=(emp) (kpts) (btex) (valid) (vvalid)](matchvalid) {};
	\end{pgfonlayer}
	\begin{pgfonlayer}{bg}
	\node[label={\small \textbf{3D Matching}}, fit=(kpts) (btex) (valid)](3dmatchvalid) {};
	\end{pgfonlayer}
	\begin{pgfonlayer}{bg}
	\node[label={\small \textbf{KF Matching}}, fit=(bow) (orbm) (vvalid)](vmatchvalid) {};
	\end{pgfonlayer}

	% Opt
	\node [blockOT, below right = -0.75cm and 1.5cm of emp2] (o1) {\small Push inter-keyframe constr. to pose graph};
	\node [blockOT, below = 0.1cm of o1] (o3) {\small Push inter-submap constr. to pose graph};
	\node [blockOT, below = 0.1cm of o3] (o4) {\small iSAM2 update};
	\begin{pgfonlayer}{bg}
	\node[draw=gray,dashed,fill=teal!5,line width=1.2pt, rounded corners, label={\small \textbf{Optimization}}, fit=(o1)  (o3) (o4)](o5) {};
	\end{pgfonlayer}

	\path [line-short] (in) -- (fe) node[pos=0.4] {};
	\path [line-sshort] (in) -- (submapgen) node[pos=0.4] {};
	\path [line-sshort] (fe) -- (submapgen) node[pos=0.4] {};
	\path [line-sshort] (subgen) -- (pq) node[pos=0.4] {};
	\path [line-sshort] (pq) -- (matchvalid);
	\path [line-sshort] (matchvalid) -- (o5);

	% Highlight contributions
	\node[line width=1pt, rounded corners, draw=red!80, inner sep=0pt, label={}, fit=(orb) ](hl1) {};
	\node[line width=1pt, rounded corners, draw=red!80, inner sep=0pt, label={}, fit=(bucket) ](hl2) {};
	\node[line width=1pt, rounded corners, draw=red!80, inner sep=0pt, label={}, fit=(msg) ](hl3) {};
	\node[line width=1pt, rounded corners, draw=red!80, inner sep=0pt, label={}, fit=(bow) ](hl4) {};
	\node[line width=1pt, rounded corners, draw=red!80, inner sep=0pt, label={}, fit=(orbm) ](hl5) {};
	\node[line width=1pt, rounded corners, draw=red!80, inner sep=0pt, label={}, fit=(pnp) ](hl6) {};
	\node[line width=1pt, rounded corners, draw=red!80, inner sep=0pt, label={}, fit=(vvalid) ](hl7) {};
	\node[line width=1pt, rounded corners, draw=red!80, inner sep=0pt, label={}, fit=(o1) ](hl8) {};

	\end{tikzpicture}
	\caption{Summary of our SLAM pipeline including parts from the previous 3D-only framework \cite{Schuster2018slam,Schuster2019phd} and the new contributions in this paper (highlighted in red): Stereo point clouds are aggregated into submaps from visual-inertial pose estimations. Submaps embed visual keyframes comprising a set of ORB descriptors as well as the camera pose relative to the submap origin (sec. \ref{sec:slam}). Once new submaps are published in the ROS network, the spatial overlap of submaps, accounting for the pose covariance, establish a priority queue of tentative matches. Matches between submaps are validated from 3D and keyframe matches, maximizing the probability of validating loop closures. 3D matches are found by clustering CSHOT descriptor matches (sec. \ref{sec::3Dmatches}) while keyframe matches are established from a set of candidates using PnP+RANSAC and subsequently validated (sec. \ref{sec::KFmatches}). All the inter-submap pose constraints resulting from validated matches are added to the graph, which is then optimized by the iSAM2 algorithm.}
	\label{fig:overview}
\end{figure*}
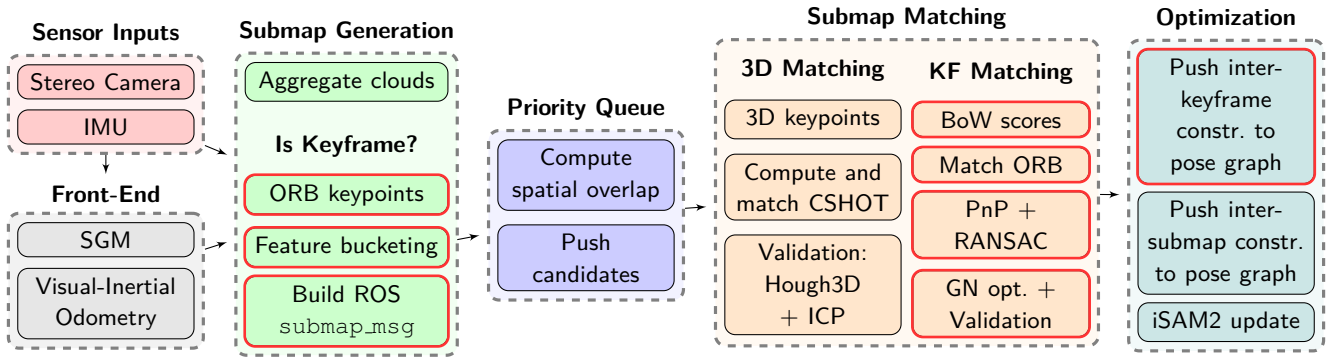
\renewcommand{\sfdefault}{phv}

\section{Related Work}\label{sec:rel}
Many works exist in the literature about loop closure in SLAM by means of visual place recognition or by performing data association with other types of sensors~\cite{Cadena2016}. In the context of mobile robotics and according to the motivation of this work, we distinguish the following categories based on whether correlations are searched amongst images or point clouds.
\subsection{Visual Place Recognition}
Visual place recognition is traditionally performed by matching local feature descriptors, such as SURF~\cite{bay2006surf}, SIFT~\cite{lowe2004distinctive}, BRISK~\cite{leutenegger2011brisk} or ORB~\cite{rublee2011orb}, usually leveraging clustering techniques to reduce the computational effort while preserving match precision~\cite{cummins2011appearance,GalvezTRO12}.
%Among these techniques, the Bag-of-Words model \cite{sivic2003video} provides a compact representation of images based on the occurrence of visual words. A vocabulary of visual features, built as kd-tree using the k-means algorithm in \cite{GalvezTRO12} allows to quickly associate visual words to descriptors and can keep track of where visual words appeared thanks to an inverted index.
Visual SLAM systems often make use of the Bag of Words (BoW) model for loop closure detection such as ORB-SLAM2~\cite{murORB2}, LDSO~\cite{gao2018ldso} or S-PTAM~\cite{pire2017s}.

A large number of works describe adaptations and improvements of the BoW model for place recognition. The authors of \cite{khan2015ibuild} and \cite{garcia2018ibow}  propose an incremental Bag of Binary Words formulation to remove the need to train a vocabulary, which is built and queried during the same session.
%In \cite{garcia2018ibow} is presented an incremental formulation which employs an efficient image indexing scheme for a fast retrieval of loop closure candidates.
In \cite{kejriwal2016high}, the authors demonstrate the benefits of exploiting the co-occurrence of pairs of visual words to increase precision.
In \cite{gehrig2017visual}, the problem of visual place recognition is efficiently solved in a probabilistic manner removing the dependency on hard thresholds and showing robustness to perceptual aliasing.
%In alternative to Bag of Word models, a Hierarchical Navigable Small World graph is used in \cite{an2019fast} to recall images based on similarity between learned features. SURF features are then used to validate image matches and compute relative transformations.
Although providing improvements over earlier works, the usual target application is for autonomous driving in urban environments, where the visual appearance is rarely ambiguous and co-occurrence of visual features is often guaranteed while revisiting the same locations. Contrarily, the absence of designated paths or tracks in completely unstructured environments does not impose any constraints on the orientation while revisiting places. 

Other approaches are targeted at improving robustness of vision-based approaches, which suffer from differences in viewpoint, non-monotonic changes in illumination and general appearance changes such as seasonal or meteorologic. The authors of \cite{burki2019vizard} propose a localization system for autonomous driving systems leveraging multiple maps that are created in different environmental conditions and merged together. This, however, requires bootstrapping initial locations using GPS. The problem of place recognition despite strong changes in viewpoint is addressed in \cite{maffra2019real} by densification of the map using local meshes to gather more feature correspondences in the context of a BoW scheme. Differently from our system, this approach relies only on visual features, which in challenging outdoor scenarios might be ambiguous.

\subsection{Point Cloud based Localization}
Other localization approaches rely on the usage of 3D Light Detection and Ranging (LiDAR) sensors, which return metrically accurate point clouds regardless of the visual appearance of the environment. Such is the case of many localization systems for autonomous driving, as common cityscapes suit well the problem of aligning 3D scans.

Detection of loop closures in this case relies on recalling similar point clouds, using global descriptors~\cite{kim2018scan} or local descriptors such as SHOT~\cite{Tombari_2011} or FPFH~\cite{rusu2009fast}. Other approaches address the problem in way similar to the visual case by analyzing similarity of depth images~\cite{steder2010narf}.

In recent years, convolutional neural networks have been used to learn more efficient local descriptors \cite{gojcic2018perfect,yew2018-3dfeatnet} that show higher performance than handcrafted approaches but require powerful hardware, which is usually unavailable in resource constrained vehicles.  A different approach is followed by the authors of SEGMAP~\cite{dube2018incremental,dube2020segmap} where a compact representation of segmented regions from point clouds are obtained through an autoencoder and matched in a different network. Despite the high performances in recalling similar places, this approach relies on the range and accuracy of 3D LiDARs and might be unsuitable for unstructured planetary environments, where it is unclear how to extract segments and 3D structure is often absent.

\subsection{Heterogeneous Approaches}
As mentioned in the review paper~\cite{piasco2018survey}, the combination of visual appearance and geometry can lead to better performances for place recognition. In~\cite{caselitz2016monocular}, cameras are localized on a 3D map built from LiDAR scans by aligning local reconstructions from a purely visual pipeline to the map. Similarly, in \cite{kim2018stereo}, stereo frames are localized with respect to a 3D map by aligning depth images. Both approaches require to bootstrap an initial pose estimate for localization. As a solution to this problem, the authors of \cite{cattaneo2019global} train a 3D and a 2D network which create a shared embedding space producing similar descriptors both for images and point clouds when they refer to the same place. In these works, localization depends on a strong correlation between visual information and structure. Contrarily, in our SLAM system, image and structure similarity are recalled independently from each other, increasing robustness for arbitrary viewpoints and trajectories.

\section{Submap-Based SLAM}\label{sec:slam}
In this section we briefly summarize the submap-based SLAM system for exploration vehicles equipped with stereo cameras introduced in our previous publications \cite{Schuster2018slam,Brand2014,Schuster2019phd}.
A schematic overview is given in Fig.~\ref{fig:overview}, highlighting the components most relevant to this paper.
Visual-inertial odometry provides locally accurate pose estimates which are used to merge local stereo point clouds into submaps. Creation of new submaps is triggered by enforcing constraints on the length of the travelled path or by the growth of pose uncertainty. Thus, submaps can be used as rigid point clouds for the purpose of loop closure and global map building. The origin of each submap is constrained by relative pose constraints and constitute nodes in a graph optimized by the iSAM2 algorithm~\cite{kaess2012isam2} from the GTSAM library.

In this work, we extend our previous point cloud based formulation of submaps to embed visual information in the form of keyframes.
While building submap $S$, the extracted keyframes are emplaced in the submap according to the pose from visual-inertial odometry relative to the submap origin. The relative position of keyframes is kept constant within each submap under the assumption that pose estimates are locally accurate.
New keyframes are saved when either the translation or rotation exceeds a threshold with respect to the last keyframes. ORB features are extracted on each new frame and bucketed across the image to achieve a uniform distribution. Depth is associated to the detected features from stereo matching computed by SGM (Semi-Global Matching~\cite{Hirschmuller2008}) running on a dedicated FPGA. Each keyframe contains the transformation from the submap origin $\mathbf{T}^s_k$ as well as its covariance $\Sigma^s_k$.

\section{Submap Matching}\label{sec:match}
 \begin{figure}[t]
	\subfloat[Original Cloud]{\includegraphics[width=0.32\linewidth]{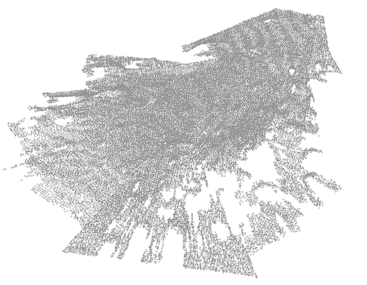}}\hfill
	\subfloat[Curvatures]{\includegraphics[width=0.32\linewidth]{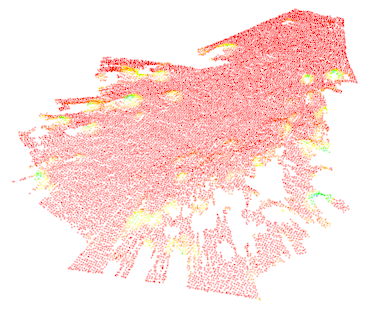}}\hfill
	\subfloat[Keypoints]{\includegraphics[width=0.32\linewidth]{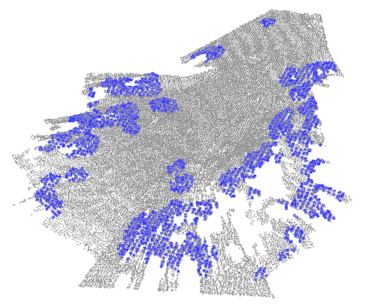}}
	\caption{Sampling keypoints on submaps from high curvature areas: After computing normals on the original submap point cloud (a), curvature is estimated from PCA (b) and points with high curvatures are used as seeds to sample 3D keypoints (c).}
        \label{fig:sampling}
\end{figure}
As a new submap is generated, loop closures are searched for by examining the overlap with older submaps, inflated according to their spatial uncertainty.
As the overlap exceeds a set score \cite{Schuster2018slam}, the corresponding pair of submaps is evaluated for possible 3D or keyframe matches.

\subsection{3D Keypoint Matching}\label{sec::3Dmatches}
Upon first evaluation for matching, 3D keypoints are extracted either on segmented obstacle regions~\cite{Brand2014} or on high curvatures regions~\cite{giubilato2020relocalization}, as illustrated in Fig.~\ref{fig:sampling}. The later approach is better suited for unstructured outdoor environments where the distinction between traversable regions and obstacles is less evident. CSHOT descriptors \cite{Tombari_2011} are matched using kd-trees on candidate submap pairs. Hough3D clustering \cite{tombari2010object} is then applied on the obtained descriptor correspondences in order to gather multiple hypotheses of transformations between the point clouds. In case, multiple hypotheses are compatible with a relative pose prior, for geometric consistency, the one with the highest number of voters is selected. The match is then refined using ICP, and a pose constraint is added to the graph.

\subsection{Keyframe Matching}\label{sec::KFmatches}
Let submaps $S_0$ and $S_1$ be candidate matches containing two sets of visual keyframes $\text{K}_0 \in S_0$ and $\text{K}_1 \in S_1$.

\vspace{0.1cm}
\noindent\textbf{BoW scoring}: The first step is to compute a BoW representation for all keyframes, storing it in a cache to avoid recomputing in case the same submap is recalled for further evaluations.
We use the DBoW2 library~\cite{GalvezTRO12} for generating BoW vectors and scoring them.
For every pair of BoW vectors $\mathbf{v}_i$ and $\mathbf{v}_j$, we compute the L1 score $s(\mathbf{v}_i$, $\mathbf{v}_j)$ proposed in~\cite{GalvezTRO12}.
For every keyframe in $\text{K}_1$ we search for the keyframe in $\text{K}_0$ that maximizes the score $s$ and such that the scores in a window of 3 keyframes centered on the candidate from $\text{K}_0$ exceed the $60\%$ of the maximum value. In fact, if adjacent keyframes in $\text{K}_0$ have comparable scores with the query in $\text{K}_1$, they probably share visual words, as if they were captured consecutively. Furthermore, we impose a global constraint on BoW scores: we keep track of the highest and lowest score observed so far ($s_{\max}$ and $s_{\min}$) and accept a candidate matching keyframe pair only if
\begin{equation}\label{eq:score_thresh}
    s(\mathbf{v}_i, \mathbf{v}_j) > 0.6\cdot(s_{\max}-s_{\min})+s_{\min},
\end{equation}
thus avoiding unnecessary evaluations. The value of the BoW score depends on a variety of factors that include properties of the vocabulary and therefore its absolute value can not be used directly. 

\noindent\textbf{Pose Estimation and Validation}: As a candidate keyframe pair is evaluated, the first step is to match all ORB descriptors, rejecting descriptor pairs whose Hamming distance is higher than 50. As mentioned previously (Sec.~\ref{sec:slam}), part of the image has depth information from disparity evaluated on a FPGA. We leverage this information to align $k_1 \in \text{K}_1$ to $k_0 \in \text{K}_0$ in a 3D-to-2D fashion, selecting from the matched feature pairs only those where depth is available in both frames. Specifically, we use the P3P algorithm \cite{gao2003complete} embedded in a RANSAC scheme to compute a tentative 6D transformation $\mathbf{T}_{k_0}^{k_1}$ using the full 3D landmarks from $k_0$ and the ORB features from $k_1$. If the number of inliers after the RANSAC test is lower than the $60\%$ of the input 3D-2D correspondences we discard the match. This value is selected from empirical considerations, in order to reject false matches but accounting for a moderate amount of outliers.  For the matches that passed the RANSAC test, we use the depth information in $k_1$ to check if the aligned sparse 3D clouds are coherent. Let $\mathbf{X}^{k_0}_j = \{x, y, z\}^{k_0}_j$ and $\mathbf{X}^{k_1}_j = \{x, y, z\}^{k_1}_j$ be two matched 3D landmarks belonging respectively to keyframes $k_0$ and $k_1$.  We therefore compute the number of ORB matched pairs such that
\begin{equation}\label{eq::depth_valid}
    |z^{k_1}_j - (R_{31}x^{k_0}_j + R_{32}y^{k_0}_j + R_{33}z^{k_0}_j + t_z)| < 0.1\,{\rm m}
\end{equation}
where $0.1$\,m is the estimated maximum depth uncertainty of the stereo camera and $j=1,2,\dots,n_{\text{inl}}$. $R_{31}, R_{32}, R_{33}$ and $t_z$ are coefficients from the bottom row of the rotation matrix $\mathbf{R}$ and translation vector $\mathbf{t}$ returned after RANSAC. Thus, \eqref{eq::depth_valid} is the difference of the z camera coordinates of an aligned pair of landmarks. If the fraction of keypoint pairs that satisfy \eqref{eq::depth_valid} is lower than the $75\%$, the match is rejected as it may indicate that the RANSAC search selected a wrong transformation which satisfied a minimum consensus from a set of few wrong keypoint matches.
This step is followed by a non-linear optimization step, refining the transformation computed after P3P-RANSAC and, furthermore, providing a covariance for the estimated pose. This optimization step aims at solving the following problem %\remWS{remove also those inliers that did not meet  \eqref{eq::depth_valid}?}:
for all feature pairs that satisfied the previous validation test \eqref{eq::depth_valid}:
%\begin{equation}\label{eq:pose_est}
%    \argmin_{\mathbf{T}_{k_0}^{k_1}} \frac{1}{2} \sum_{j=1}^{n_\text{}} ||\mathbf{x}_j - \pi(\mathbf{X}_j^{k_0}, \mathbf{T}_{k_0}^{k_1})||^2
%\end{equation}
\begin{equation}\label{eq:pose_est}
    \argmin_{\mathbf{T}_{k_0}^{k_1}} \sum_{j=1}^{n} \rho(r_j) \qquad r_j = |\mathbf{x}_j - \pi(\mathbf{X}_j^{k_0}, \mathbf{T}_{k_0}^{k_1})|
\end{equation}
where $\rho()$ refers to the Cauchy robust loss function \cite{barron2019general},
%\remWS{use $\mathbf{u}_j$ instead of $\mathbf{x}_j$ and $\mathbf{x}_j^{k_0}$ instead of $\mathbf{X}_j^{k_0}$?}
$\mathbf{x}_j$ denotes the location of the keypoint $j$ in the image of $k_1$, $\mathbf{X}_j^{k_0}$ denotes the 3D coordinates of landmark $j$ in the reference frame of $k_0$ and $\pi$ is the projection function to the image of $k_1$. After solving \eqref{eq:pose_est}, we obtain the covariance of the pose
%\begin{equation}\label{eq::pose_cov}
%    \Sigma(\mathbf{T}_{k_0}^{k_1}) = (J(\mathbf{T}_{k_0}^{k_1})^T \ \Sigma_x^{-1} \  J(\mathbf{T}_{k_0}^{k_1}))^{-1}
%\end{equation}
by extracting the marginal covariance related to the pose parameters. Equation~\eqref{eq:pose_est} is solved with the Gauss-Newton method using the GTSAM library.

Finally, before accepting a keyframe match, we compute the transformation between the submap origins induced by the keyframe match using~\eqref{eq::tfs}.
Let $s_0$ and $s_1$ be the origins of two submaps containing the matching keyframes $k_0$ and $k_1$ respectively. Given the transformation between keyframes $\mathbf{T}_{k_1}^{k_0}$ and visual-inertial constraints, the transformation between submap origins is defined by
\begin{eqnarray}\label{eq::tfs}
\mathbf{T}_{s_0}^{s_1} & = & \mathbf{T}_{k_1}^{s_1} \ \mathbf{T}^{k_1}_{k_0} \ \mathbf{T}_{s_0}^{k_0} \\ \nonumber
& = & \mathbf{T}_{k_1}^{s_1} \ \mathbf{T}^{k_1}_{k_0} \ (\mathbf{T}_{k_0}^{s_0})^{-1} \nonumber
\end{eqnarray}
where $\mathbf{T}_{k_0}^{s_0}$ and $\mathbf{T}_{k_1}^{s_1}$ are poses from visual-inertial odometry with respect to each submap origin and $\mathbf{T}_{k_0}^{k_1}$ results from the keyframe match.
As the IMU makes the roll and pitch angles observable and submaps are gravity-aligned as provided, we verify that the roll and pitch components of $\mathbf{T}_{k_0}^{k_1}$ are close to zero or negligible. If a keyframe match satisfies all these checks, it is used to add a constraint to the graph as described in the following.
\vspace{0.1cm}

\noindent
\textbf{Loop closure constraints}:
%Let $s_0$ and $s_1$ be the origins of two submaps containing the matching keyframes $k_0$ and $k_1$ respectively. Given the transformation between keyframes $\mathbf{T}_{k_1}^{k_0}$ and visual-inertial constraints, the transformation between submap origins is defined by
%\begin{eqnarray}\label{eq::tfs}
%\mathbf{T}_{s_0}^{s_1} & = & \mathbf{T}_{k_1}^{s_1} \ \mathbf{T}^{k_1}_{k_0} \ \mathbf{T}_{s_0}^{k_0} \\ \nonumber
%& = & \mathbf{T}_{k_1}^{s_1} \ \mathbf{T}^{k_1}_{k_0} \ (\mathbf{T}_{k_0}^{s_0})^{-1} \nonumber
%\end{eqnarray}
%where $\mathbf{T}_{k_0}^{s_0}$ and $\mathbf{T}_{k_1}^{s_1}$ are poses from visual-inertial odometry with respect to each submap origin and $\mathbf{T}_{k_0}^{k_1}$ results from the keyframe match.
%Each transformation is affected by errors and the uncertainty is modeled as a zero-mean Gaussian noise $\mathcal{N}(0, \Sigma)$, see Fig.~\ref{fig:kf_match} for a graphical overview.
%Equation \eqref{eq::tfs} therefore is rewritten as
%\begin{equation}\label{eq::tf_wcov}
%[\mathbf{T}|\Sigma]_{s_0}^{s_1} = [\mathbf{T}|\Sigma]_{k_1}^{s_1} \ [\mathbf{T}|\Sigma]_{k_0}^{k_1} \ [\mathbf{T}|\Sigma]^{k_0}_{s_0}
%\end{equation}
%\remWS{Looks like a simple matrix multiplication -- true also for $\Sigma$s?}
\begin{figure}[t]
    \centering
    \includesvg[width=\linewidth]{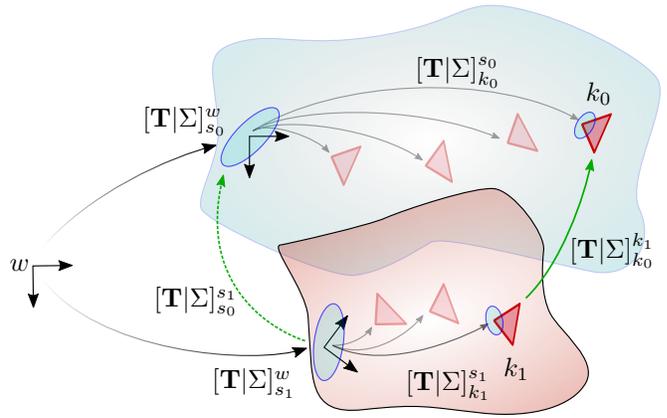}
    \caption{Overview of the transformations involved in the process of submap matching where $k_0$ and $k_1$ are matching keyframes belonging respectively to submaps $s_0$ and $s_1$. Green arrows denote matching constraints: the solid line denotes inter-keyframe constraints (visual matches, Sec.~\ref{sec::KFmatches}) while the dotted line denotes inter-submap constraints (3D matches, Sec.~\ref{sec::3Dmatches}). Black arrows denote instead constraints from visual-inertial odometry: from world to submap origins and from submap origins to keyframes.}
    \label{fig:kf_match}
\end{figure}
%The resulting transformation between submap origins constrains the relative pose in a pose graph, in addition to possible additional constraints from submap matches based on 3D correspondences (Sec.~\ref{sec::3Dmatches}).
%The covariance associated to the inter-submap pose constraint ~\eqref{eq::tfs} is derived by composing the intermediate uncertain transformations.
%Following the notation in \cite{mangelson2019characterizing}, let $p_0$, $p_1$ and $p_2$ be three arbitrary 6D poses and let $f_\oplus$ and $f_\ominus$ denote the operations of pose composition and inversion such that
%\begin{eqnarray}
%	\mathbf{T}_{p_2}^{p_0}&=&f_\oplus(\mathbf{T}_{p_1}^{p_0}, \mathbf{T}_{p_2}^{p_1}) = \mathbf{T}_{p_1}^{p_0} \mathbf{T}_{p_2}^{p_1} \\ \nonumber
%	\mathbf{T}_{p_1}^{p_0}&=&f_\ominus(\mathbf{T}_{p_0}^{p_1}) = (\mathbf{T}_{p_0}^{p_1})^{-1} \nonumber
%\end{eqnarray}
%Being $J_\oplus$ and $J_\ominus$ the Jacobians for the operations of pose composition and inversion and $p_0$, $p_1$ and $p_2$ arbitrary poses, we use the operations %:
%\begin{eqnarray}
%	\Sigma_{p_2}^{p_0} & = & J_\oplus(\mathbf{T}_{p_1}^{p_0}, \mathbf{T}_{p_2}^{p_1}) \ \hat{\Sigma} \ J_\oplus(\mathbf{T}_{p_1}^{p_0}, \mathbf{T}_{p_2}^{p_1})^T \\ \nonumber
%	\Sigma_{p_1}^{p_0} & = & J_\ominus(\mathbf{T}_{p_0}^{p_1}) \ \Sigma_{p_0}^{p_1} \ J_\ominus(\mathbf{T}_{p_0}^{p_1})^T \nonumber
%\end{eqnarray}
%to combine $[\mathbf{T}|\Sigma]^{s_1}_{k_1}$ with $[\mathbf{T}|\Sigma]^{k_1}_{k_0}$ and the resulting transformation with the inverse of $[\mathbf{T}|\Sigma]^{s_0}_{k_0}$.
Once a match is validated between keyframes $k_0$ and $k_1$, the respective poses are added as nodes to the graph (Sec.~\ref{sec:slam}) and are constrained with respect to the origins of the parent submaps with $[\mathbf{T}|\Sigma]^{s_0}_{k_0}$ and $[\mathbf{T}|\Sigma]^{s_1}_{k_1}$ (see Fig~\ref{fig:kf_match}). The pose and covariance $[\mathbf{T}|\Sigma]^{k_1}_{k_0}$ instead constrain the global poses of the matching keyframes. In addition, eventual submap matches based on 3D correspondences (Sec.~\ref{sec::3Dmatches}) establish a constraint between the origins of submaps $S_0$ and $S_1$. All inter-submap and inter-keyframe constraints are added to the graph using the Cauchy robust error model \cite{lee2013robust}, as implemented in the GTSAM library, for robustness against outliers from any eventual false constraint. This eventuality is very unlikely given the sequence of validation checks. Contrarily to the multi-robot or multi-session case, where it is beneficial to gather as many candidate loop closures and discriminate between them for consistency (e.g. switchable constraints \cite{sunderhauf2012switchable} or PCM \cite{mangelson2018pairwise}), we make use of the strong prior on the robot pose, weighted by its uncertainty, to discard in advance erroneus matches.

\section{Experiments}\label{sec:exp}
In this section we describe tests of our SLAM system in two different environments demonstrating the benefit of using multiple modalities for matching submaps built from stereo camera measurements. The test sites were a laboratory environment at the DLR Institute of Robotics and Mechatronics, Germany, and a designated planetary analog environment on Mount Etna, Italy~\cite{wedler2015robex,wedler2017robex,EtnaDatasets}.

\subsection{Indoor Sequences}
\begin{figure}[t]
\centering
\subfloat[\textit{rmc\_lab\_long} top view]{\includegraphics[width=\linewidth]{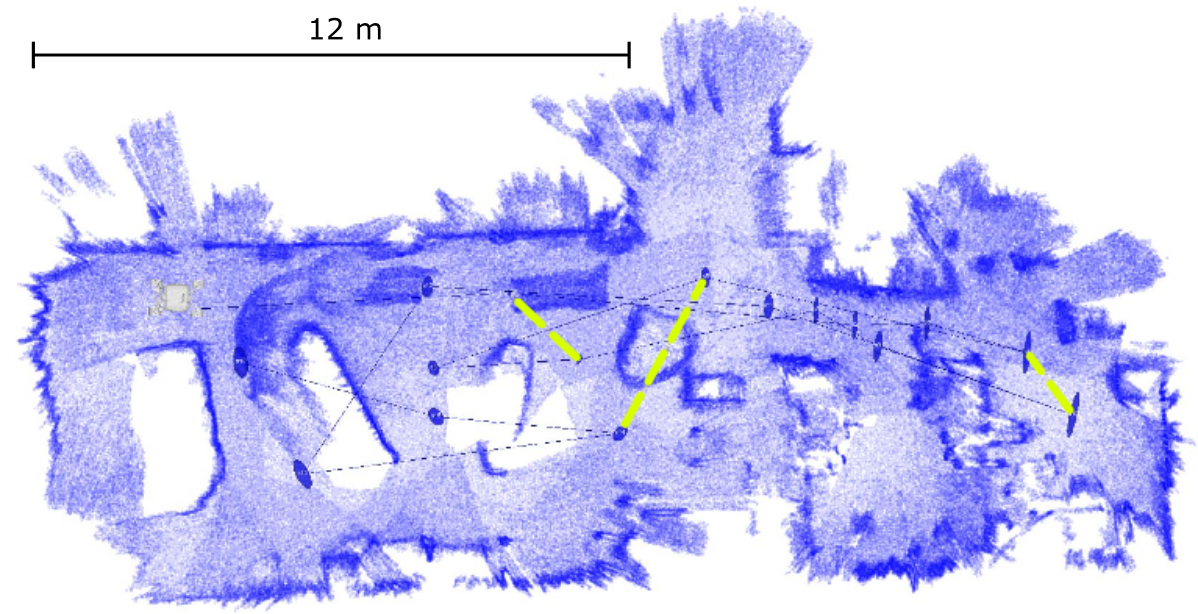}} \\
%\subfloat[\textit{rmc\_corr} detail]{\includegraphics[width=0.6\linewidth]{fig/corr_zoom.png}\label{fig:lab:detail}} \hfill \subfloat[\textit{rmc\_corr} full top view]{\includegraphics[width=0.4\linewidth]{fig/corr_full.png}}
\subfloat[\textit{rmc\_corr} full top view]{\includegraphics[width=0.4\linewidth]{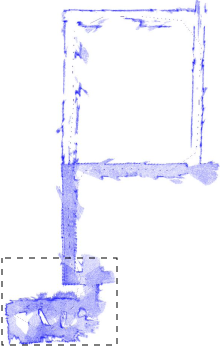}}\hfill\subfloat[\textit{rmc\_corr} detail]{\includegraphics[width=0.6\linewidth]{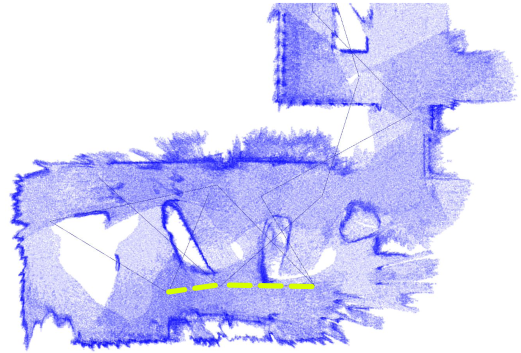}\label{fig:lab:detail}}
\caption{Top views of the \textit{rmc\_lab\_long}~(a) and the \textit{rmc\_corr}~(b) sequences; (c)~shows of detail of \textit{rmc\_corr} as indicated by dashed gray rectangle in~(b). Dashed yellow lines highlight the inter-submaps constraints after establishing loop closures. The accuracy of pose estimates after SLAM updates can be visually inferred from the coherence of the point clouds in the proximity of the walls and rocks in the lab.}\label{fig:lab}
\end{figure}
\begin{table}[t]
\caption{RMSE errors in the RMC sequences with multiple matching strategies. Lowest error in bold, highest in brackets}
\centering\label{table::lab}
\renewcommand*\arraystretch{1.3}
\begin{tabular}{|c|l|cccc|}\hline
\textbf{Sequence} & \textbf{RMSE}     & \textbf{No Matches} & \textbf{3D} & \textbf{KF} & \textbf{3D+KF} \\
	     \hline\hline
 \multirow{4}{*}{\textit{rmc\_lab\_long}}      & pos [m]          & 0.47  & {(0.54)} & 0.27 & \textbf{0.21} \\ \cline{2-6}
\ & z [m]            & {(0.20)}  & 0.08 & 0.05 & \textbf{0.08} \\ \cline{2-6}
& angle [$^\circ$] & {(4.18)}  & 1.47 & 0.91 & \textbf{0.90} \\ \cline{2-6}
& $n_\text{matches}$& - & 4 & 2 & \textbf{5} \\ \hline\hline
\multirow{4}{*}{\textit{rmc\_corr}}    & pos [m]          & {(1.31)}   & 0.24 & 0.94 & \textbf{0.20} \\ \cline{2-6}
& z [m]            & {(0.25)}  & 0.09 & 0.20 & \textbf{0.09} \\ \cline{2-6}
& angle [$^\circ$] & \textbf{0.81}  & 2.18 & {(4.07)} & 0.86 \\ \cline{2-6}
& $n_\text{matches}$& - & 4 & 1 & \textbf{5} \\ \hline\hline
\multirow{4}{*}{\textit{rmc\_lab}}    & pos [m]                             & {(0.51)} & 0.33 & 0.31 & \textbf{0.28} \\ \cline{2-6}
& z [m]            & {(0.15)} & \textbf{0.09} & \textbf{0.09} & \textbf{0.09} \\ \cline{2-6}
& angle [$^\circ$]                    & {(2.68)} & 2.56 & \textbf{1.65} & 2.48 \\ \cline{2-6}
& $n_\text{matches}$& - & 3 & 1 & 3 \\ \hline
\end{tabular}
\end{table}
The first test environment is an indoor laboratory featuring a number of large stone replicas which provide the look of a rocky outdoor environments.
The laboratory features a ceiling mounted Vicon tracking system which provides absolute poses (rotation and translation) for an accurate ground truth with a coverage of about~$70\,\text{m}^2$.
In these sequences, the obstacles, or non-traversable regions, provide unique 3D structures where the extracted CSHOT descriptors should be unambiguous.
As the obstacles are easily distinguishable from the remaining environment (walls and floor), we extract 3D keypoints on segmented obstacles from depth images \cite{Schuster2018slam,Brand2014}.

To evaluate the quality of pose estimation, we compute position and angular errors given the difference between Vicon poses and submap origins, which are the one optimized by iSAM2. Let $\mathbf{T}^w_{s_i}$ and $\hat{\mathbf{T}}^w_{s_i}$ be the estimated and true transformations from a global reference system to the origin of the $i$-th submap. Thus, the transformation $\mathbf{T}_{\text{err}} = \mathbf{T}^w_{s_i} \ (\hat{\mathbf{T}}^w_{s_i})^{-1}$ describes the difference between the true and estimated pose of submap $i$. Being $\mathbf{T}_{\text{err}} = [\mathbf{R}_{\text{err}} \ | \ \mathbf{t}_{\text{err}}]$, we estimate the position error using the L2 norm of the translation part $\|\mathbf{t}_{\text{err}}\|^2$ and the orientation accuracy by computing the total angle from the rotation term $\mathbf{R}_{\text{err}}$ as
\begin{equation}
	\phi_{\text{err}} = \arccos\Big(\frac{\text{Tr}(\mathbf{R}_{\text{err}})-1}{2}\Big)
\end{equation}
where $\text{Tr}(\mathbf{R}_{\text{err}})$ is the trace of the rotation matrix.

The three sequences, \textit{rmc\_lab\_long}, \textit{rmc\_corr} and \textit{rmc\_lab} always start and end inside the laboratory, visible in Fig.~\ref{fig:lab:detail}, such that ground truth is always available at the beginning and end of the trajectory.
Table~\ref{table::lab}
reports the errors evaluated in all the indoor sequences.
We test our SLAM system enabling the 3D and keyframe matching (KF) modules independently to observe which configuration leads to higher performances.
We also report the errors obtained from the visual-inertial odometry which, as expected, are the highest.
Overall, the combination of the two matching strategies leads to the best performances as the chance of validating submap matches is higher.
While in our previous works \cite{Schuster2018slam,giubilato2020relocalization} the presence of unique 3D structures was necessary, in the proposed framework, submaps can match either from similarity in structure, visual appearance or both.
In the 3D+KF case, the number of matched submaps is usually higher, as is consequently the number of constraints pushed to the optimization graph. In the laboratory datasets, furthermore, the number of matched submaps using the 3D strategy is usually higher than those using keyframes.
The extent of each submap make them comprise many 3D structures which can easily be matched, while
keyframe matches need the camera to return very close to previously visited places, as the stereo head is usually tilted towards the ground in order to observe the presence of obstacles.
%Nevertheless, if many keyframe matches appear in a single submap pair, the pose constraint from keyframes only is very accurate.
Figure~\ref{fig:lab} shows the top views of two laboratory sequences highlighting the connection between submaps from odometry and from map matches, using both 3D structure and keyframes. Examples of submap matches with matches from keyframes and 3D descriptors highlighted in different colors are presented in Figures~\ref{fig:sub_16_match1} and \ref{fig:sub_16_match2}.

\subsection{Outdoor Sequences}
\begin{figure}[t]
\centering
\subfloat[Aligned trajectories with DGPS ground truth]{\includesvg[width=1\linewidth]{fig/run9_rev}\label{fig:etna_plot}}  \\
\subfloat[Example image]{\includegraphics[width=0.48\linewidth, trim=2.5cm 2cm 2cm 2cm, clip]{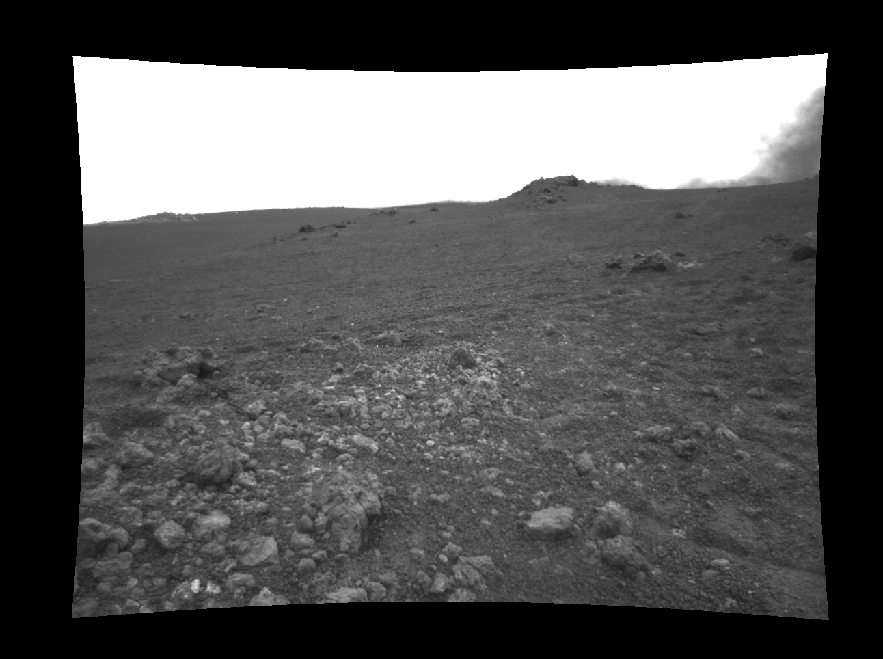}\label{fig:etna_cam}} \hfill
\subfloat[Inter-submap constraints]{\includegraphics[width=0.48\linewidth]{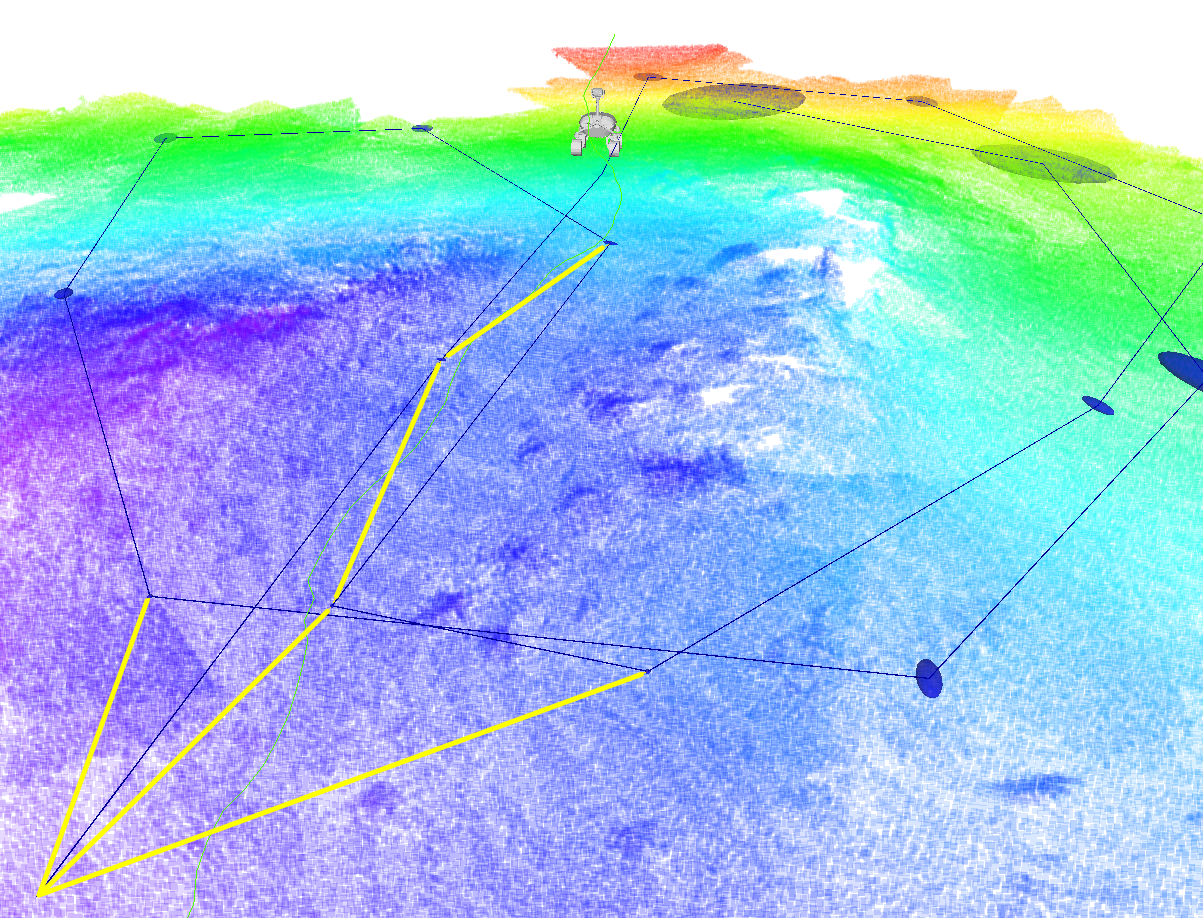}\label{fig:etna_graph}}
\caption{Full trajectories from the Etna sequence aligned to the DGPS ground truth with Loop Closure enabled and disabled. The green path corresponds to the KF+3D map matching strategy, which is the most accurate according to Table \ref{table:etna}. The magenta path is estimated by ORB-SLAM2 before a tracking failure (magenta cross). Squares denote the positions where relocalization occurred for ORB-SLAM2, however followed by other tracking failures. The orange path is estimated by RTAB-MAP. (b) Example of camera image from this Etna sequence. (c) Detail of map and pose graph with inter-submap constraints after matching (yellow lines).}\label{fig:etna9}
\end{figure}
\begin{table}[t]
\caption{Pose errors after alignment with the DGPS ground truth in the Etna sequence for multiple map matching strategies}\label{table:etna}
\centering
\renewcommand*\arraystretch{1.3}
\begin{tabular}{|l|cccc|}\hline
	     & \textbf{No Matches} & \textbf{3D only} & \textbf{KF only} & \textbf{3D+KF} \\
	     \hline\hline
RMSE [m] & {(0.38)}       & 0.32 & 0.30 & \textbf{0.24} \\ \hline
$n_\text{matches}$ & - & 2 & 4 & \textbf{5} \\ \hline
\end{tabular}
\end{table}

\begin{figure}
	\subfloat[]{\includegraphics[width=0.5\linewidth, trim=1cm 1cm 1cm 1cm, clip]{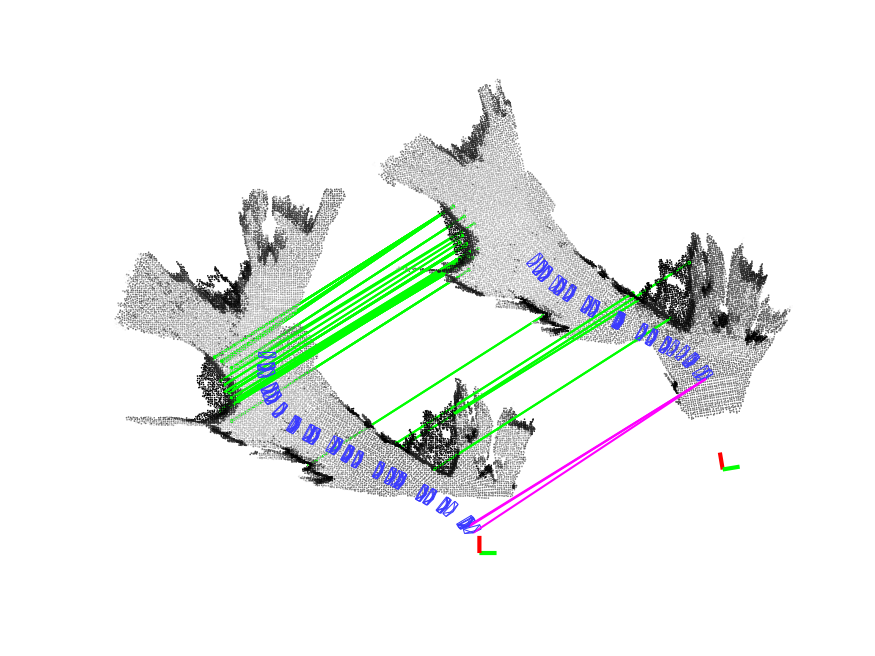}\label{fig:sub_16_match1}}\hfill
	\subfloat[]{\includegraphics[width=0.5\linewidth, trim=1cm 1cm 1cm 1cm, clip]{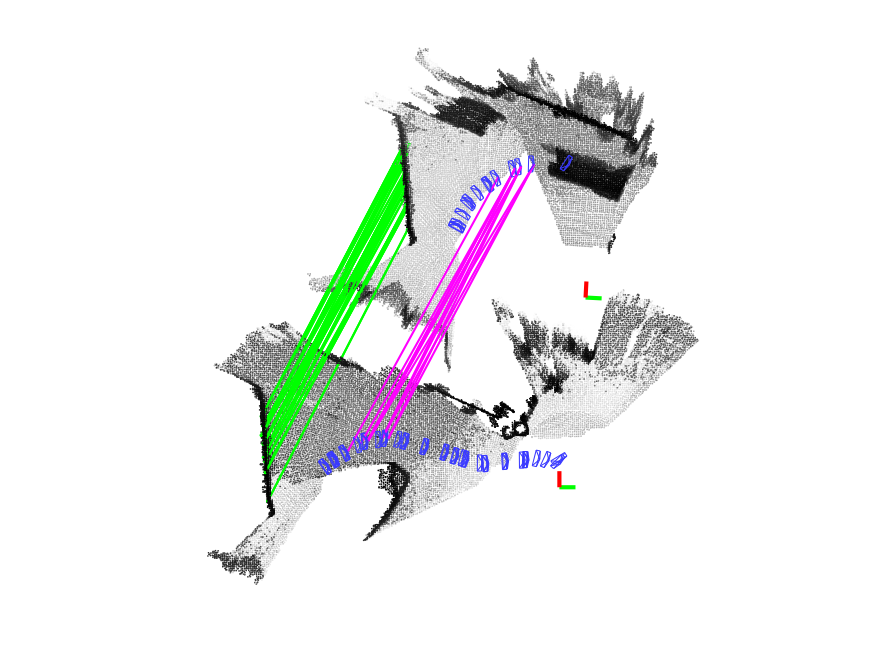}\label{fig:sub_16_match2}} \\
	\subfloat[]{\includegraphics[width=0.5\linewidth, trim=1cm 1cm 1cm 1cm, clip]{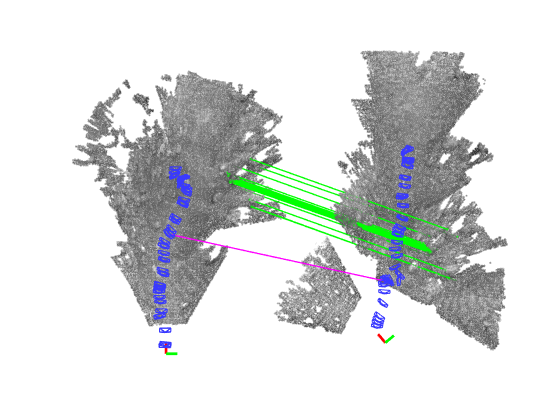}\label{fig:sub_etna_match1}}\hfill
	\subfloat[]{\includegraphics[width=0.5\linewidth, trim=1cm 1cm 1cm 1cm, clip]{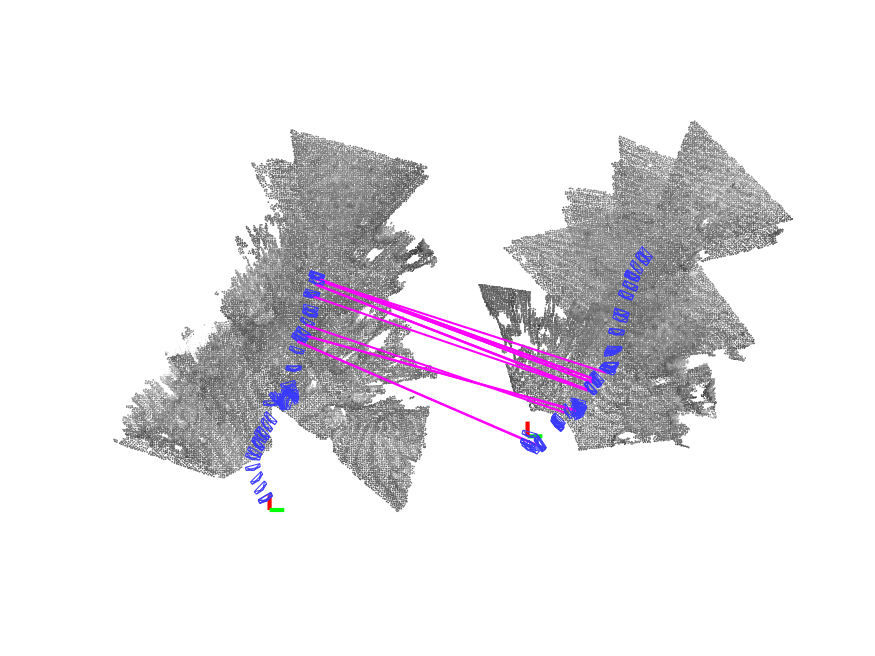}\label{fig:sub_etna_match2}}
	\caption{Heterogeneous submap matches: green lines are 3D keypoint matches from CSHOT correspondences, magenta lines are keyframe matches from ORB descriptor correspondences. In (b) the submap match is established only by keyframe correspondences, while in (a) and (c) 3D matches are predominant. (a) and (b) are submap matches from the Etna sequence, while (c) and (d) are example matches from \textit{rmc\_lab\_long}. For all submaps, the lenght of the trajectory is approximately 7 meters.}
\end{figure}
In this section we evaluate the performances of our pipeline on a sequence recorded on Mount Etna, Italy.
The moon-like environment includes a distribution of small volcanic rocks resulting in an ambiguous visual appearance (see Fig.~\ref{fig:etna_cam} for an example camera view).
A DGPS setup provides global position measurements as ground truth.
Thus, contrarily to the indoor sequences, we evaluate here only position errors after aligning the SLAM results to the DGPS track given correspondences between the timestamps.
Poses are aligned to the ground truth using Horn's quaternion method~\cite{horn1987closed} optimizing rotations and fixing the relative pose from the first timestamp correspondence. With respect to Fig.~\ref{fig:etna_plot}, the LRU rover drives on two circular paths: a first one clockwise in the bottom right and a second one counterclockwise on the top left. Two major loop closure opportunities are located first on the middle traverse, where the rover revisits the previously travelled path from the same direction, and second corresponding to a crossing between trajectories in the area of coordinates $(-2, 4)$.

Figure~\ref{fig:etna9} shows the aligned trajectories estimated from our pipeline and computed with loop closure disabled and enabled with the 3D+KF matching strategy. The trajectories in blue and green are plotted by transforming local trajectories relative to each submap into world coordinates, using the transformations from submaps to world as estimated by the most recent iSAM2 update.
The plot in Fig.~\ref{fig:etna_plot} shows that the SLAM results after loop closures better fit the ground truth, while the trajectory from visual-inertial odometry-only accumulates drift that occur especially during rotations.
In fact, the RMSE pose errors reported in Table~\ref{table:etna} with loop closure enabled decrease in the order RMSE(3D-only)$>$RMSE(KF-only)$>$RMSE(KF+3D).
This is because the combination of different modalities can establish a larger number of pose constraints that are also more widely distributed along the graph, see Fig.~\ref{fig:etna_graph} for an example of loop closures. Figures~\ref{fig:sub_etna_match1} and \ref{fig:sub_etna_match2} show two examples of submap pairs from the Etna sequence where matches are obtained predominantly either with 3D descriptors (Fig.~\ref{fig:sub_etna_match1}) or with keyframes (Fig.~\ref{fig:sub_etna_match2}), motivating the need for combining different modalities in such challenging scenarios.
Note that, with respect to Table~\ref{table:etna}, the improvement of a single submap match is quite significant: submaps span areas of approximately 7 meters in length and therefore represent large portions of the mapping session.  

To compare the performance of our system with other state of the art SLAM approaches, we tested ORB-SLAM2 as well as RTAB-MAP \cite{labbe2019rtab} in RGB-D configuration using the recorded depth from SGM, computed on an FPGA onboard the rover. As can be seen in Fig.~\ref{fig:etna9}, the trajectory estimated by ORB-SLAM2 stopped after about 7 meters from the start as a result of tracking failures due to the repetitive and ambiguous appearance of the challenging moon-like scenario. Note the presence of two successful relocalizations (magenta boxes) followed by an immediate tracking failure. The trajectory estimated by RTAB-MAP, as seen marked in orange in Fig.~\ref{fig:etna9}, successfully completes the first half of the sequence and stops near the end as a result of tracking failure. RTAB-MAP is able to detect a loop closure when the rover revisits the first traverse but fails to detect the second one. As a result, the first half of the trajectory is quite accurate and comparable with the one estimated from our pipeline but the drift accumulated in the second half is not compensated, see the top-left part of Fig.~\ref{fig:etna9}.

\begin{figure}[t]
    \captionsetup[subfigure]{labelformat=empty}
    \subfloat[]{\includegraphics[width=\linewidth]{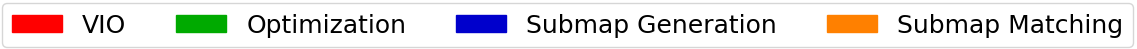}}\\\vspace{-0.5cm}
    \subfloat[]{\includegraphics[width=\linewidth]{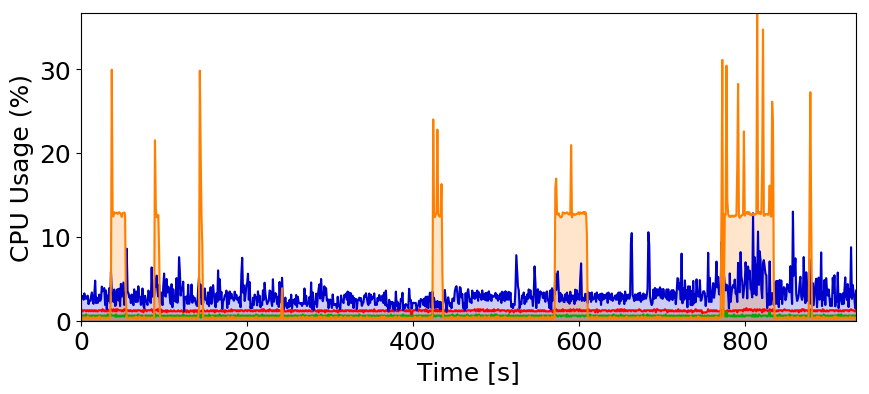}}
    \caption{CPU usage of the relevant parts of the SLAM pipeline as introduced in Fig.~\ref{fig:overview} (100\% corresponds to all 8 CPU cores).}
    \label{fig::perf}
\end{figure}

Fig.~\ref{fig::perf} shows the CPU load introduced from the main components of the SLAM pipeline. Both the visual-inertial odometry as well as the graph updates require almost constant CPU time. The submap generation task involves not only accumulating stereo point clouds into the current submap but also incorporating keyframes and extracting ORB features. The most expensive task is submap matching due to the high dimensionality of C-SHOT descriptors. The time spent on keyframe matching is, infact, a marginal part of this thanks to the binary ORB features and the BoW-based candidate selection. Overall, Fig.~\ref{fig::perf} shows that the presented SLAM pipeline has minimal impact on the rover's onboard computer which is important to allow other software components (e.g. path planning, state machines, etc.) to operate smoothly.

\section{Conclusions}\label{sec:concl}
In this work, we presented a novel approach to establishing loop closures in a submap-based SLAM system by leveraging structure and appearance similarity. Candidate submap matches, selected from prior knowledge of the robot pose and uncertainty, are validated by searching correspondences both across 3D keypoint descriptors and visual keyframes. We tested the proposed system both in indoor laboratory environments with replicas of natural structures as well as in a sequence captured in a planetary analogous scenario on Mount Etna, Italy, showing the benefits in terms of pose accuracy given by matching submaps in different modalities. Future developments of this pipeline involve
evaluating alternative terrain representations to pointclouds to overcome noise and occlusions \cite{le2020gaussian}, investigating segmentation techniques for natural landmarks \cite{chiodini2020evaluation} by jointly using 3D and visual information and, finally, 
extending the existing framework to a multi-robot system comprising both ground and aerial vehicles \cite{muller2018robust}.

\section*{ACKNOWLEDGMENT}
\label{sec:acknowledgement}
This work was supported by the Helmholtz Association, project alliance ROBEX (contract number HA-304) and project ARCHES (contract number ZT-0033).

\bibliographystyle{IEEEtran}
\bibliography{biblio}

\end{document}